\documentclass{article}
\usepackage{ijcai25}

\usepackage{times}
\usepackage{soul}
\usepackage{url}
\usepackage[hidelinks]{hyperref}
\usepackage[utf8]{inputenc}
\usepackage[small]{caption}
\usepackage{graphicx}
\usepackage{amsmath}
\usepackage{amsthm}
\usepackage{booktabs}
\usepackage{algorithm}
\usepackage{algorithmic}
\usepackage[switch]{lineno}

\usepackage[T1]{fontenc}

\usepackage{listings}
\usepackage{xcolor}
\usepackage{amsmath}

\title{Let's Measure the Elephant in the Room: Facilitating Personalized Automated Analysis of Privacy Policies at Scale}

\author{Rui Zhao \and
Vladyslav Melnychuk \and
Jun Zhao \and Jesse Wright \and Nigel Shadbolt
\affiliations
University of Oxford, Oxford, UK\\
\emails
\{rui.zhao,jun.zhao\}@cs.ox.ac.uk, 
vladyslav.melnychuk18@gmail.com, 
\{jesse.wright,nigel.shadbolt\}@jesus.ox.ac.uk
}

\definecolor{codegreen}{rgb}{0,0.6,0}
\definecolor{codegray}{rgb}{0.5,0.5,0.5}
\definecolor{codepurple}{rgb}{0.58,0,0.82}
\definecolor{backcolour}{rgb}{0.95,0.95,0.92}

\lstdefinestyle{upstyle}{
    backgroundcolor=\color{backcolour},   
    commentstyle=\color{codegreen},
    keywordstyle=\color{magenta},
    numberstyle=\tiny\color{codegray},
    stringstyle=\color{codepurple},
    basicstyle=\ttfamily\footnotesize,
    breakatwhitespace=false,         
    breaklines=true,                 
    captionpos=b,                    
    keepspaces=true,                 
    numbers=left,                    
    numbersep=5pt,                  
    showspaces=false,                
    showstringspaces=false,
    showtabs=false,                  
    tabsize=2
}

\newcommand{\ws}[1]{\lstinline|#1|}
\newcommand{\up}[1]{\lstinline|#1|}

\begin{document}

\maketitle

\begin{abstract}
% A small fraction of Web users read the Terms of Service or Privacy Policies of online platforms, despite being requisite for service usage.
In modern times, people have numerous online accounts, but they rarely read the Terms of Service or Privacy Policy of those sites despite claiming otherwise.
This paper introduces PoliAnalyzer, a neuro-symbolic system that assists users with personalized privacy policy analysis. PoliAnalyzer uses Natural Language Processing (NLP) to extract formal representations of data usage practices from policy texts.
% Using logical inference, PoliAnalyzer then compares user preferences with the formal privacy policy representation to produce a compliance report.
In favor of deterministic, logical inference is applied to compare user preferences with the formal privacy policy representation and produce a compliance report. 
To achieve this, we extend an existing formal Data Terms of Use policy language to model privacy policies as \textit{app policies} and user preferences as \textit{data policies}.
In our evaluation using our enriched PolicyIE dataset curated by legal experts, PoliAnalyzer demonstrated high accuracy in identifying relevant data usage practices, achieving F1-score of 90-100\% across most tasks. Additionally, we demonstrate how PoliAnalyzer can model diverse user data-sharing preferences, derived from prior research as 23 user profiles, and perform compliance analysis against the top 100 most-visited websites. This analysis revealed that, on average, 95.2\% of a privacy policy’s segments do not conflict with the analyzed user preferences, enabling users to concentrate on understanding the 4.8\% (636 / 13205) that violates preferences, significantly reducing cognitive burden. Further, we identified common practices in privacy policies that violate user expectations - such as the sharing of location data with 3rd parties. 
This paper demonstrates that PoliAnalyzer can support automated \textit{personalized} privacy policy analysis at scale using off-the-shelf NLP tools. This sheds light on a pathway to help individuals regain control over their data and encourage societal discussions on platform data practices to promote a fairer power dynamic.

\end{abstract}

\section{Introduction}

Collectively, we are subscribing to an ever-increasing number of online services - each of which have us sign custom ``Terms of Service'' or ``Privacy Policies'' to enable the collection and use of our data. 
% Modern life made us have numerous accounts on different online platforms, gathering and storing our data and activities, each under an agreement between us and the platform under Terms of Service or Privacy Policy. 
Despite the privacy and legal implications, less than 7\%~\cite{obar_biggest_2020} of consumers read these agreements making them the ``the biggest lie on the Internet.'' 
% However, despite the legal implications, it is generally known that people seldom read such terms, aka. ``the biggest lie on the Internet'' \cite{obar_biggest_2020,robinson_beyond_2020}. 
\citeauthor{obar_biggest_2020} suggest the cause is information overload, with the average privacy policy requiring a 29-minute read-time. 
% Research \cite{obar_biggest_2020} has suggested that a major cause for this situation is information overloading due to the length and the number of policies, which is not easy to overcome given current legal and technical infrastructure.
Current regulation, including General Data Protection Regulation (GDPR)~\cite{council_of_european_union_regulation_2016} and the Digital Service Act (DSA)~\cite{council_of_european_union_regulation_2022} exacerbate the issue, requiring companies to collect more permissions from users without providing technical or legal standards for facilitating users.

% Legal regulations like General Data Protection Regulation (GDPR) \cite{council_of_european_union_regulation_2016} or Digital Service Act (DSA) \cite{council_of_european_union_regulation_2022} will not resolve this issue of users' reading efforts, as platforms will still create their own heterogeneous policies.

% Indeed, there are increasing legal attention to user privacy, data usage, and accountability, such as the European Unions's General Data Protection Regulation (GDPR) \cite{council_of_european_union_regulation_2016}, Digital Service Act (DSA) \cite{council_of_european_union_regulation_2022}, Artificial Intelligence Act (AI Act) \cite{council_of_european_union_regulation_2024}. However, it is unrealistic to expect legal documents to enforce all platforms to use a uniformed data usage practice, thus saving users from (expected to) reading such policy documents.

% This creates a restless situation, especially considering the increasing discussion of user agency and autonomy in digital world \cite{wu_slow_2023,stein_you_2023,calacci_access_2023}, which often expects users to spend more time and focus in controlling their digital presence or the services they use. Connected with the increasing call and practice for decentralization and decentralized governance \cite{meurisch_privacy-preserving_2020,chen_decentralized_2021,gehl_digital_2023,anaobi_will_2023}, the neglected issue of not able to read such policies will become stronger.
% To some extent, it has become an elephant in the room -- everyone knows it, but there is limited discussion on dealing with it.

One solution is to require that online privacy policies have a formal, uniform or interoperable, machine-readable presentation; this enables user agents - such as browsers or dedicated systems - to parse them and warn users of any terms violating their policies preferences. Research on this topic often falls under the theme of data usage control~\cite{zhao_perennial_2024,breaux_eddy_2014,sandhu_usage_2003} or legal modeling and reasoning~\cite{robaldo_reified_2017,prakken_law_2015,palmirani_pronto_2018}, with different focuses and solutions.
% A line of research provides a potential direction to improve the situation, by utilizing computer-interpretable formal representations of data usage policies, and performing automated reasoning to verify compliance between the specified policy and the actual data usage context. Such research often falls under the theme of data usage control \cite{zhao_perennial_2024,breaux_eddy_2014,sandhu_usage_2003} or legal modelling and reasoning \cite{robaldo_reified_2017,prakken_law_2015,palmirani_pronto_2018}, where each model has different focuses and thus different strategies and solutions. 

Given the absence of formal representations published by platforms, it is essential to explore automated methods for constructing or generating policy encodings. This work seeks to address the following question: can personalized analysis of privacy policies be facilitated at scale using existing resources?

% As websites do not yet publish such formal representations, it would be valuable to generate them from the existing natural text.
% However, an important question is not yet well-explored: how to construct the formal policy encoding, so we can use them for reasoning and compliance checking?
% In this paper, we assess whether these representations can be generated at scale, and the computational efficency of personalized compliance checks.
% whether personalized compliance checks can be applied efficiently.
% In this paper we address the questions: ``Can these representations be generated at scale?'' and ``Can personalized compliance checks be applied at scale?''

% In this work, we aim to bridge one facet of this gap and provide more. In particular, we aim to answer this question: is it possible to design a system that facilitates personalized analysis of privacy policies at scale?

Recent advances in Large Language Models (LLMs) are transforming the field of NLP. LLMs have advanced performance on language-understanding tasks~\cite{wang_superglue_2019,hendrycks_measuring_2020,srivastava_beyond_2023}, and can be adapted to new tasks through few-shot prompting~\cite{brown_language_2020,wang_generalizing_2020,liu_pre-train_2023} or fine-tuning with minimal data~\cite{sanh_multitask_2021,chung_scaling_2022}.
% We see the opportunity from recent advances of natural-language-processing (NLP) domain, represented by large language models (LLMs). LLMs have shown advanced performance in different language-understanding tasks \cite{wang_superglue_2019,hendrycks_measuring_2020,srivastava_beyond_2023}, and have been shown to be adaptable to different tasks through the means of few-shot prompting/learning \cite{brown_language_2020,wang_generalizing_2020,liu_pre-train_2023} or fine-tuning without significant data \cite{sanh_multitask_2021,chung_scaling_2022}. 
These advances provide opportunities for citizens to extract formal descriptions from privacy policy text with minimal effort.

% This provides an exciting potential to identify and extract relevant information about data usage practices from privacy policies with minimal effort from regular users, which can then be converted into formal policy encodings.
% Our work is related to but differs from existing approaches in using NLP tools to analyze privacy policies (Section \ref{sec:background:pp-analysis}) given our different focus and downstream tasks.

By having LLMs generate a formal description, we allow logical inference engines to then compare user preferences against policies rather than LLMs. This mitigates potential logic errors and explainability challenges~\cite{kaur_trustworthy_2022,dwivedi_explainable_2023,zhao_explainability_2024} that would come with using an LLM for this compliance checking task.

% rather than directly compare user-preferences to privacy policies we eliminate the need 
% Formal policy descriptions provide a precise and consistent way of representing concepts from a range of policy documents. They enable explainable evaluation of user preferences against policies, eliminating possible errors and explainability challenges~\cite{kaur_trustworthy_2022,dwivedi_explainable_2023,zhao_explainability_2024} that would come with using an LLM for this task.

% Using formal policy encodings provides two benefits: 1) the policy encoding as logical rules is accurate and interoperable, and can be easily applied in a different context, such as a different type of documents; 2) it circumvents an important limitation of LLMs in their explainability \cite{kaur_trustworthy_2022,dwivedi_explainable_2023,zhao_explainability_2024}, as formal reasoning can be audited by verifying the logical proofs, while LLMs are used as a proxy for annotators, rather than reasoners or judges.

% Choosing an appropriate formal policy language is also important, because different policy language exhibit different features and properties (Section \ref{sec:background:dtou-policy}).

Putting this together, we present the PoliAnalyzer system, which uses state-of-the-art LLMs to perform information extraction from privacy policies, and converts the extracted information about data practices into formal \textit{app policies} of an extended version of the psDToU (Perennial Semantic Data Terms of Use) \cite{zhao_perennial_2024} policy language. PoliAnalyzer can also consult a logical reasoner to check the compliance status between the constructed \textit{app policies} and users' preferences encoded as \textit{data policies} to perform personalized analysis, of different scales.

The remaining sections follow this structure: in Sec \ref{sec:background}, we review existing literature, and discuss the distinctions of our work; in Sec \ref{sec:system}, we present the design and design insights of PoliAnalyzer; Sec \ref{sec:evaluation:nlp-pipeline} presents our evaluation of the NLP pipeline, as a measurement of underlying technology feasibility; Sec \ref{sec:evaluation:platform-user} forms an evaluation of data practices of contemporary online world through the lens of PoliAnalyzer with real-life user expectations, also demonstrating how PoliAnalyzer can support the interests of different personnel in performing automated analysis of privacy policies at scale.

\paragraph*{Contribution}

This paper makes the following main contributions:
\begin{enumerate}
   \item \textit{We provide the first toolkit for generating formal data usage policies from privacy policy texts}, filling in a gap in current data governance and usage policy research.
   % It fills in a gap of the current data governance and usage policy research, by utilizing (off-the-shelf) state-of-the-art NLP technology to construct formal data usage policy from widely-existed privacy policy documents;
   \item \textit{We provide the first system for automated \textbf{personalized} analysis of privacy policies}, as a pathway to improve online privacy, transparency and user-agency.
   \item \textit{We assess whether the top 100 leading online platforms meet real-life users' expectations for how their data is used}, identifying that on average 4.8\% (636 / 13205) of the policy segments violate user expectations commonly mentioned in existing research;
   \item \textit{We identify common privacy policy practices that violate user expectations}, such as location data being shared with 3rd parties;
   \item \textit{We perform comprehensive evaluation of off-the-shelf LLMs for complex privacy policy queries}, through an enriched privacy policy dataset, demonstrating their appropriateness, and identifying challenges.

   % By enriching privacy policy annotation corpus, we evaluated the NLP pipeline to validate its feasibility, and identified challenges for state-of-the-art NLP technologies in this context;
   % \item We introduce the first system enabling automated \textbf{personalized} analysis of privacy policies of online platforms, PoliAnalyzer, which opens a route to improve user agency, and a new paradigm of online privacy and transparency discussion;
   % \item Based on PoliAnalyzer, we evaluated how well the top 100 leading online platforms meet real-life users' expectations for how their data to be used, also demonstrating its ability to reduce user attention by 94.7\%.
\end{enumerate}

\section{Background and Related Work}
\label{sec:background}

\subsection{Privacy Policy Analysis}
\label{sec:background:pp-analysis}

Existing work on privacy policy analysis spans across several themes, and we focus on those on identifying and re-presenting privacy policy information, and supporting users decision-makings.

To assist comprehension of privacy policies, works including Tos;DR (Terms of Service; Didn't Read)\footnote{https://tosdr.org/} and privacy (nutrition) labels or icons \cite{kelley_nutrition_2009,emami-naeini_ask_2020,efroni_privacy_2019} suggest using a fixed set of icons that highlight key policy information. These icons are manually created by developers (e.g.~App Store's nutrition label) or crowdsourcing (e.g.~ToS;DR). Despite simple and easy to learn, these approaches are not expressive enough to capture a large portion of policy terms, and do not support personalized policy analysis.
% Work like Tos;DR (Terms of Service; Didn't Read)\footnote{https://tosdr.org/} and privacy (nutrition) labels or icons \cite{kelley_nutrition_2009,emami-naeini_ask_2020,efroni_privacy_2019} 
% suggested using simple limited number of indications like icons to summarize the important information, either manually created by the developers (e.g.~App Store's nutrition label) or crowdsourcing (e.g.~ToS;DR). However, these approaches make it hard to capture custom information requirements or personalization.

% Zaeem, et al. \cite{zaeem_privacycheck_2018} proposed the PrivacyCheck system to automatically identify and summarize several aspects of privacy policy to facilitate users.
% Some research \cite{liu_have_2021,hamdani_combined_2021,xiang_policychecker_2023} also used natural-language processing (NLP) technologies to automatically identify information from privacy policies and check that against GDPR requirements. They expanded the generality of information identification, but the focus on GDPR made them less useful for users who are not legal workers.

% \cite{liu_have_2021} proposed the AutoCompliance system to automatically analyze mobile Apps' privacy policies with regard to their compliance status of GDPR (Article 13), about how the privacy policy (not) meets the legal requirements.
% Likewise, \cite{hamdani_combined_2021} proposed a NLP-based approach to automataically checking the compliance of privacy policies of GDPR requirements.
% \cite{xiang_policychecker_2023} proposed

Some work proposed self-trained NLP models to identify certain types of information from privacy policies, such as PolicyLint \cite{andow_policylint_2019}, Polisis \cite{harkous_polisis_2018} and PoliGraph \cite{cui_poligraph_2023}, with support of downstream tasks such as converting to privacy icons, supporting custom queries, and internal consistency analysis. However, it is unclear how regular users can make use of these tools, given the required familiarity of information schema of their internal representation, nor for personalized analysis. In addition, because the models are not for general tasks, the user is also required to set up the technical environment themselves, increasing potential burden.

% PoliCheck \cite{andow_actions_2020} investigates the \emph{behavior} of apps, where privacy policy is used to calibrate the data flow detection. 
% PolicyLint \cite{andow_policylint_2019} uses NLP tools to identify information from PPs, and convert that to a formal representation, to study the contradictions within the PPs.
% Polisis \cite{harkous_polisis_2018} developed deep-learning-based NLP tools for PP information extraction, and demonstrated how the extracted information can be used to create privacy icons and help free-form queries.
% PoliGraph \cite{cui_poligraph_2023} proposed the use of knowledge graph to facilitate the privacy policy analysis, and trained NLP models for the information extraction.

There is also prior work in using 
% Apart from custom models, some work explored the possibility of using 
off-the-shelf LLMs to analyze legal documents or privacy policies. In particular, \cite{savelka_unreasonable_2023}, LegalBench \cite{guha_legalbench_2023}, PolicyGPT \cite{tang_policygpt_2023} and \cite{rodriguez_large_2024} evaluated different LLMs' performances against certain types of queries using existing datasets. Such research showed the advantages of using off-the-shelf LLMs for the privacy policy annotation tasks, thus incentivizing our design choice. However, they face the challenge of query tasks being simple, such as simply asking for boolean answers to the existence of different types of data practices, which is not enough for our downstream task.

% Apart from using custom models, some work explored the possibility of using off-the-shelf LLMs to analyze legal documents or privacy policies. \cite{savelka_unreasonable_2023} evaluated the possibility of using LLMs of the GPT family to annotate legal documents, in zero-shot settings. LegalBench \cite{guha_legalbench_2023} is a comprehensive benchmark for legal-related tasks, and utilized OPP-115 and APP-350 datasets. However, they only used simple question-answering, and lacks detailed comparison with ground truth values. PolicyGPT \cite{tang_policygpt_2023} evaluated ChatGPT, GPT-4 and Claude2 in their respective performance in classifying data practices in OPP-115 and PPGDPR \cite{liu_have_2021}. \cite{rodriguez_large_2024} compared LLama 2 \cite{touvron_llama_2023} and GPT-4 (Turbo) in their respective performance on asking types of data being collected (from a given list of data types), of MAPP \cite{arora_tale_2022} and OPP-115 dataset, using different prompting strategies and fine-tuning. Such research showed advantages of using off-the-shelf LLMs for the privacy policy annotation tasks, thus incentivized our design choice; however, they did not cover all tasks required in our work.

% PLUE \cite{chi_plue_2023} benahmarked several privacy policy datasets using non-llms, and reported performance. In general, their performance is worse than ours. Although, their intent classification is better than ours.

Compared with existing work, this paper has three distinct features: 1. the analysis is personalized rather than general; 2. the result should be explainable, for easier auditing; 3. the method should be accessible by regular users. Reflected in Figure \ref{fig:arch-polianalyzer} below, it forms improvements in flow (a) by evaluating LLM performance in complex tasks, and provides additional flows, (b) and (c).

% Overall, our work is between the intersection of prior work in extracting information from PPs (such as \cite{harkous_polisis_2018,cui_poligraph_2023}) and supporting decision-making (such as \cite{zaeem_privacycheck_2018,liu_have_2021}). But we have some distinct expectations that are not covered in existing research: 1. the analysis should be personalized rather than general, saving user's focus; 2. the result should be explainable, for easier auditing. We also need to evaluate the performance of LLMs to verify their utility in our specified tasks.

\subsection{Privacy Policy Corpus}
\label{sec:background:pp-corpus}

Previous work has created several datasets for privacy policies, with different focus.
Notably, the Usage Privacy Policy project \cite{sadeh_usable_2013} released multiple datasets, especially OPP-115, APP-350 and Privacy QA as to be described below.

OPP-115 \cite{wilson_creation_2016} is a well-known early annotation dataset for 115 website privacy policies, and is widely used by later research. It contains annotations of nine types of data practices, each with further detailed questions, and the span of texts. Each annotation is performed on a given ``paragraph-length'' policy segment.
Likewise, APP-350 \cite{zimmeck_maps_2019} is a dataset of 350 mobile app privacy policies, focusing on the data type, the party, and modality.

PI-Extract \cite{bui_automated_2021} presented a 30-document dataset containing data types and practices (collecting or not, sharing or not). PrivacyQA \cite{ravichander_question_2019} (35 documents) and PolicyQA \cite{ahmad_policyqa_2020} (built on OPP-115) focused more on natural-language question-answering, where they collected questions related to privacy or policy concerns.

Policy-IE \cite{ahmad_intent_2021} is a distinct dataset that contains more detailed information about data practices, of 31 documents.
In addition to the data practices, it also contains granular information for them, especially party, action, (data and purpose) entity, and the \textit{relation} between them and the practices.

Given the granularity of information, we chose the Policy-IE dataset as it contains the most comprehensive information for our needs. In particular, the relations information in the dataset is crucial for identifying which role each different type of entities constitutes in a data practice.
However, this dataset still lacks formal, or unified, names for different types of entities, which is why we enriched the model in this work (see Section \ref{sec:evaluation:nlp-pipeline}).

\section{PoliAnalyzer}
\label{sec:system}

\begin{figure}
   \centering
   \includegraphics[width=\linewidth]{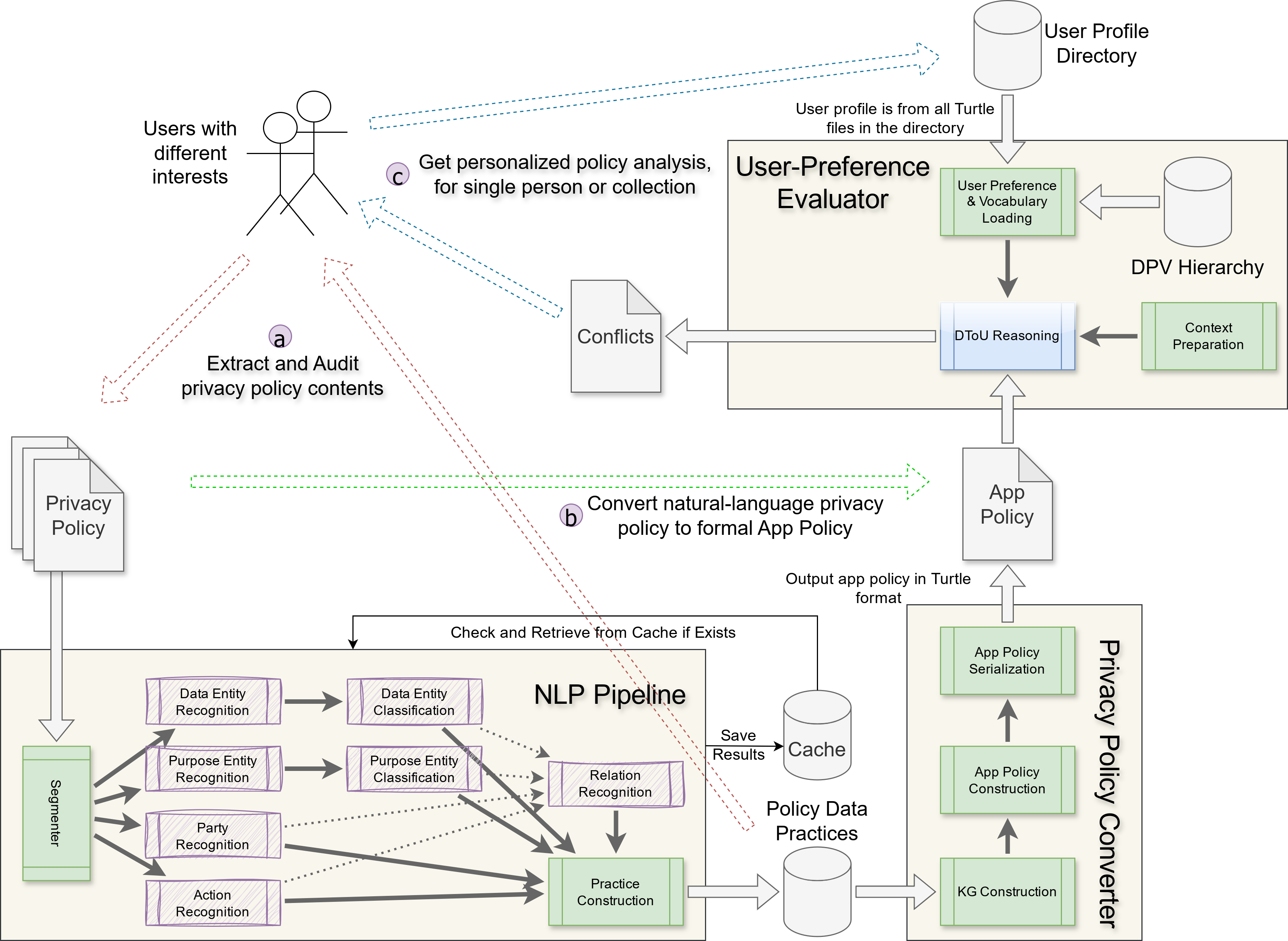}
   \caption{Design of PoliAnalyzer. Solid arrows represent data flow within PoliAnalyzer; dashed arrows represent main user flows.}
   \label{fig:arch-polianalyzer}
\end{figure}

We introduce PoliAnalyzer, which supports the previously described functionalities, facilitating Web users to perform personalized, automated analysis of privacy policies. It consists of three main components: the NLP pipeline, the privacy policy converter, and the user-preference evaluator, as shown in Figure \ref{fig:arch-polianalyzer}. The supplementary material contains the code of PoliAnalyzer, the LLM prompts, and examples of formal policy encoding. This section describes the main component design and usage.

\subsection{NLP pipeline}

The \textit{NLP pipeline} is responsible for identifying relevant information from the privacy policy, where the types of information are based on PolicyIE with additions (Sec \ref{sec:evaluation:nlp-pipeline}). The flow in Figure \ref{fig:arch-polianalyzer} demonstrates the relation between individual steps within the NLP pipeline.

% \begin{itemize}
%    \item Data entities: both the data entity span of text, and the data category as in DPV;
%    \item Purpose entities: both the purpose entity span of text, and the purpose category as in DPV;
%    \item Party entities: who the party is, among \textit{user}, \textit{first party}, and \textit{third party};
%    \item Data practice: what data practices the segment talks about, among (first-party or third-party) data collection, third-party data sharing, data storage retention, and data security protection.
%    \item Relations: how the previously identified types of information relate to each other, especially what entities belong to what data practice, and serve what role (e.g.~is a party the data provider or data consumer).
% \end{itemize}

% The types of information being identified is derived from the annotation schema of PolicyIE dataset, as well as our additional annotations, to be discussed in Section \ref{sec:evaluation:nlp-pipeline}.

Each step is performed as one query of the LLM\footnote{For our prototype, we use the gpt-4o family of models, but the system design is generic to all models.}, to a segment of privacy policy, to identify the relevant information. We use each \emph{line} of the privacy policy as a segment, as a balance between accuracy and cost, based on piloting empirical experiments. For both data entity and purpose entity, they both first undergo an entity identification step (aka.~named entity recognition \cite{li_survey_2022}), and then an entity classification step of the recognized entity. Finally, all entities and data practices are grouped together by segment, and each segment is sent to perform relation identification. The results from these steps are then used to construct the data practices.

For data and purpose classification, the model will output the grounded terms in DPV, for better interoperability in later components. Specifically, because purposes have a hierarchy, the model should predict the most accurate leaf node (subclass) in the hierarchy.

When invoking LLM predictions, in general, we use system prompt to instruct the model to perform a specific task with task description, and output JSON description; we choose hyperparameters to get stable output (esp.~\lstinline|temperature = 0|); we then provide the segment to analyze from user prompt. We fine-tuned the models for each task, with a small portion of our data to reinforce both the tasks and the output schema. As an exception, for relation identification, we give each entity and practice a unique ID from our code, and send them together in user prompt.
Because LLMs sometimes do not follow the instruction, the output is not always in the expected form (which also forms the reason to perform fine-tune). We perform some post-processing before parsing the results, following heuristics discovered from response data, and existing helper libraries, especially \texttt{json-repair} \cite{jong_josdejongjsonrepair_2024}.

\subsection{Privacy Policy Converter}

The \textit{privacy policy converter} parses the results from NLP pipeline, and constructs structured representations of them. In particular, it converts the results into an internal knowledge graph, and also the structured \textit{app policy}.

\subsubsection{Formal policy mapping}

The target \textit{app policy} representation is based on the schema from psDToU (perennial semantic Data Terms of Use) policy language \cite{zhao_perennial_2024}, which can represent both the application policy and users' data policy in one model for automated compliance analysis.
For our context, each platform is considered as one application, and data usage practices are encoded as \emph{input specifications}. Most often, each \emph{input specification} (\lstinline{:input_spec}) describes what data is taken as input (\lstinline|:data|) from this \textit{port} (\lstinline|:port|, a unique identifier) within the application, whether and what third parties will receive the data for processing (\lstinline|:downstream|), and the purposes of the data processing (\lstinline|:purpose|).

% To convert the results into \textit{app policy}, we need to define a mapping from the identified information to the target policy language.

% In this work, we chose to represent our app policy using the perennial semantic Data Terms of Use (PSDToU) policy language \cite{zhao_perennial_2024}, which can represent both the application policy and users' data policy in one model. The \textit{app policy} describes application's data usage activities.

% Specifically for our context, each platform is considered as one application, and we are most interested in the \emph{input specification} construct, which organizes data practices as \emph{ports}. Each port is associated with one \emph{input specification} (\lstinline{:input_spec}), describing what data is taken as input (\lstinline|:data|) from this port within the application, whether and what third parties will receive the data for processing (\lstinline|:downstream|), and the purposes of the data processing (\lstinline|:purpose|).

Specifically, we map \texttt{collection-use} practices to \emph{input specification}, covering the data being used (as \lstinline|:data|) and the purpose being used (as \lstinline|:purpose|); we map all \texttt{third-party-sharing-disclosure} practices to \emph{downstream} of the same data, with their corresponding purposes and users (third-party name).

% In addition, the input specification has an \emph{action} (\lstinline|:action|) field, which, in our context, is used to denote whether the port processes (for collection-use and third-party-sharing-disclosure practices) or stores (for storage-retention practices) the data.

%% Maybe add an example? Or in appendix?

\subsection{User-preference evaluator}

To check whether the \textit{app policy} is compliant with users' preferences, we introduce the \emph{user preference evaluator}. The user preferences are represented as \textit{data policies} in psDToU, containing information such as what is permitted and prohibited for data usage; multiple \textit{data policies} for different data types form a \emph{user profile}, covering a user's full preferences.

Since we use DPV for data types and purposes, the reasoner can automatically identify the equivalence and relation between those specified in \textit{app policies} and \textit{data policies}. Specifically, because \emph{purposes} have hierarchy, we extended the language and reasoner to allow \textit{data policies} to specify how to match against a hierarchy tree, such as all subclasses or exact matches. Likewise, the user can also define custom hierarchies, simply by defining the sub-class-of relation for relevant entities in RDF, allowing for the representation of new concepts while fitting into existing concepts.

The formal reasoner, based on Notation3 (N3) \cite{berners-lee_n3logic_2008}, takes all policies as input, and produces compliance analysis results, containing what conflicts exist, and their details such as the original policy texts. That is presented to the user for their further usage.

\section{NLP Pipeline Evaluation}
\label{sec:evaluation:nlp-pipeline}

As discussed earlier in Section \ref{sec:background:pp-corpus}, existing work on using LLM with PP mainly focused on different main categories (data practices and data entity recognition), with little discussions about supporting the representation of relations. We conducted relevant experiments to fill in this gap, focusing on evaluating whether LLMs are appropriate for PoliAnalyzers' targeted jobs.

\begin{figure*}
   \centering
   \includegraphics[width=.9\linewidth]{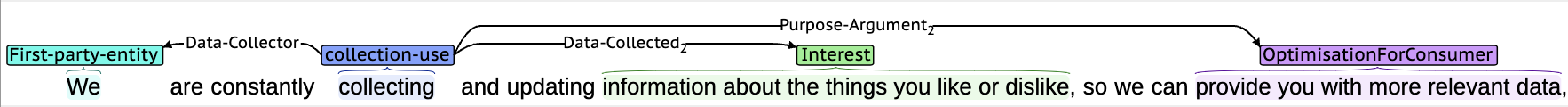}
   \caption{Sample of annotation produced by annotator.}
   \label{fig:annotation-sample}
\end{figure*}

Since no existing dataset satisfies our goal (Sec \ref{sec:background:pp-corpus}), to support the evaluation, we enriched the Policy-IE dataset, by employing two domain experts to perform additional annotations, especially focused on assigning canonical labels from DPV \cite{data_privacy_vocabularies_and_controls_community_group_data_2024} to data entities and purpose entities; they were also asked to perform other tasks, such as separating the practices which were logically two distinct practices despite mentioned in the same sentence with the same action word. Figure \ref{fig:annotation-sample} presents a sample annotation from our dataset. Please refer to the supplementary material for details on the annotation procedure and results.

With the enriched annotation, we performed a series of experiments to evaluate the NLP pipeline's performance. After comparing the performance between different prompting strategies, we chose to fine-tune two off-the-shelf models, gpt-4o and gpt-4o-mini\footnote{More specifically, we use model gpt-4o-2024-08-06 and gpt-4o-mini-2024-07-18, which were the latest at the point of experiment.}. Specifically, we tested their performance (f1-score) using the enriched dataset, where a small portion of the dataset was used as training data (120 out of 1087 data points\footnote{Note we intentionally limited the portion of data used for training, contrary to usual practices when training new models, due to our different goal: to evaluate existing LLMs.}). There are two parts of reasons for performing fine-tuning with the small portion of data: 1) the LLM output should comply with our specified schema (which is otherwise before fine-tuning); and 2) the LLM should learn some simple preferences hard to describe easily through instructions.

\begin{table*}[h]
    \centering
    \includegraphics[width=\linewidth]{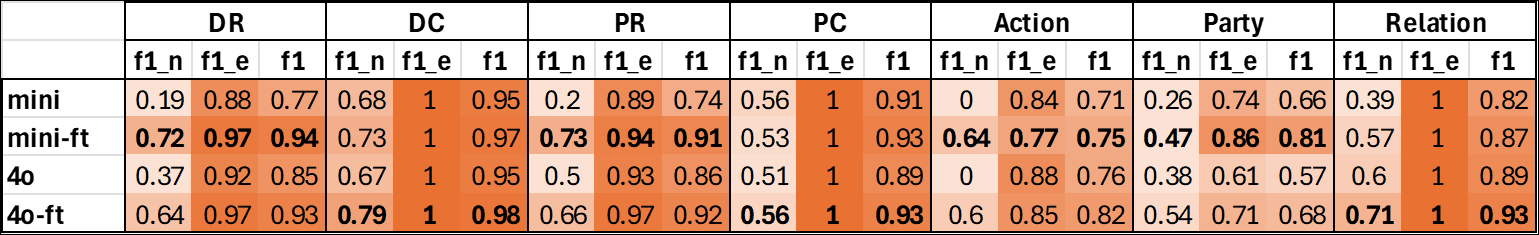}
    \caption{LLM model performance in evaluation. DR means Data Recognition; DC means Data Classification; PR means Purpose Recognition; PC means Purpose Classification; Action, Party and Relation means the corresponding recognition job. f1\_n refers to f1-non-empty metric; f1\_e refers to f1-empty metric; f1 refers to macro f1 metric. \texttt{rx} means \emph{relaxed} matching.}
    \label{tab:llm-eval-result}
\end{table*}

Table \ref{tab:llm-eval-result} summarizes the main results of the best performant models from our experiments, where the bold texts indicate the models we chose in our final system. For tasks with word matching, we use a relaxed metric where the predicted result is proportionately considered in scoring -- if the longest-common-substring ($lcs$) ratio of the (predicted and expected) result is over a given threshold ($0.9$), the true positive is increased by $lcs$; otherwise, only exact matching is considered true positive.

Overall, for the best-performant models, most tasks have f1-score of about 0.9 or higher. This means the models are able to correctly identify the desired entities in most tasks. In particular, the very high f1-empty scores indicate that the models are particularly good at determining whether the given segment contains targeting information or not. This means that the model will unlikely produce non-existent entities for the privacy policy, reducing the worry about hallucination.

On the other hand, when focusing on the scores for only non-empty-valued segments (f1-n), the score becomes lower, to around 0.5 - 0.7. This indicates that the models are not always accurate in predicting the entities, and still have space to improve. Having said that, the performance of the models does not indicate they are unacceptably worse than human annotators, as the inter-annotator agreements are also not perfect (e.g.~that for data types is 85\% for the final phase in our dataset), and the model performance is, despite lower, reasonably comparable to that.

We also notice that gpt-4o is not necessarily better than gpt-4o-mini, nor fine-tuning is always better, such as from that in purpose recognition and purpose classification. Also, sometimes the model was already good at the tasks (esp.~data and purpose classification) without fine-tuning. This potentially indicates that detailed and targeted prompts in job descriptions may already be enough for instructing the model.

In general, our result shows evidence that state-of-the-art LLMs are able to produce meaningful predictions for identifying information for data practices with small fine-tuning, reaching performance comparable to human annotators. Therefore, they can be used to annotate privacy policies, and can be utilized in PoliAnalyzer to perform large-scale analysis.

\section{Platform Policy Evaluation}
\label{sec:evaluation:platform-user}

The previous section validated our design, and this section describes our evaluation of platform policies given the feature of PoliAnalyzer. In particular, we use PoliAnalyzer to evaluate how well online platforms can actually meet \textbf{users' expectations} in data usages. We focus on the top 100 most visited websites, based on the Tranco list \cite{le_pochat_tranco_2019}\footnote{We use the Tranco list \texttt{93VV2}, resembling most-visited websites between 3 July - 1 August 2024.}. We used the privacy policy dataset from the Princeton-Leuven Longitudinal Privacy Policy Dataset \cite{amos_privacy_2021}.\footnote{Because some websites do not have valid privacy policies in the dataset, we retry the next one until 100 policies are retrieved.}

\subsection{Evaluation design}

As a high-level overview, we used PoliAnalyzer to first convert the platforms' privacy policies into \emph{app policies}; we also synthesized real-world user expectations as different user profiles, encoded as \emph{data policies}; then we performed conflict analysis between each pair of the policies, through the remaining of PoliAnalyzer. Further, we drew conclusions from the analysis procedure and results.

To the best of our knowledge, there is no dataset or structured description of real-world user preferences in data usage practices. Therefore, we derive user preferences by reviewing existing literature on user preference of data usage 
(\cite{lee_privacy_2017,benisch_capturing_2011,lin_modeling_2014,middleton_global_2020,wilson_privacy_2013}), to identify common requirements. It is worth noting that this effort is not intended as a systematic review, but provides a quick review of the latest discussions about users' data preferences in the existing literature. This provides a source of user data preferences identified in previous research, which helps us to evaluate the capability of our system.
As a summary, we identified 23 distinct data types discussed in the above literature, only 15 of which are represented in DPV; 11 purpose types, 8 of which are in DPV; and 14 types of practices discussed or highlighted in them.
In the end, we constructed 23 data policy sets, covering 7 types of data, 6 distinct purposes and 2 different types of data consumers, as summarized in Table \ref{tab:user_preference_factors}. To accommodate certain requirements in the user profiles, we extended the psDToU policy language. Details of the modelled user profiles can be found in the supplementary material, as well as an example with explanation.

\begin{table}[h]
   \small
   \centering
   \begin{tabular}{|c|c|}
       \hline
       Data Types & \begin{tabular}{@{}c@{}}
            SocialCommunication, Contact, \\
            Data-general, MedicalHealth, Identifying, \\
            Location, Picture
       \end{tabular} \\
       \hline
       Purposes & \begin{tabular}{@{}c@{}}
             Internal, Advertisement, Analytics,\\
             Research, SNS, ProtectionOfPublicSecurity
       \end{tabular} \\
       \hline
       Data Consumers & 1st-party-only, 1st-and-3rd-party \\
       \hline
   \end{tabular}
   \caption{Factors in the user preferences}
   \label{tab:user_preference_factors}
\end{table}

\subsection{Result \& Discussion}
\label{sec:exp-results:platform-user}

\begin{figure}
   \centering
   \includegraphics[width=\linewidth]{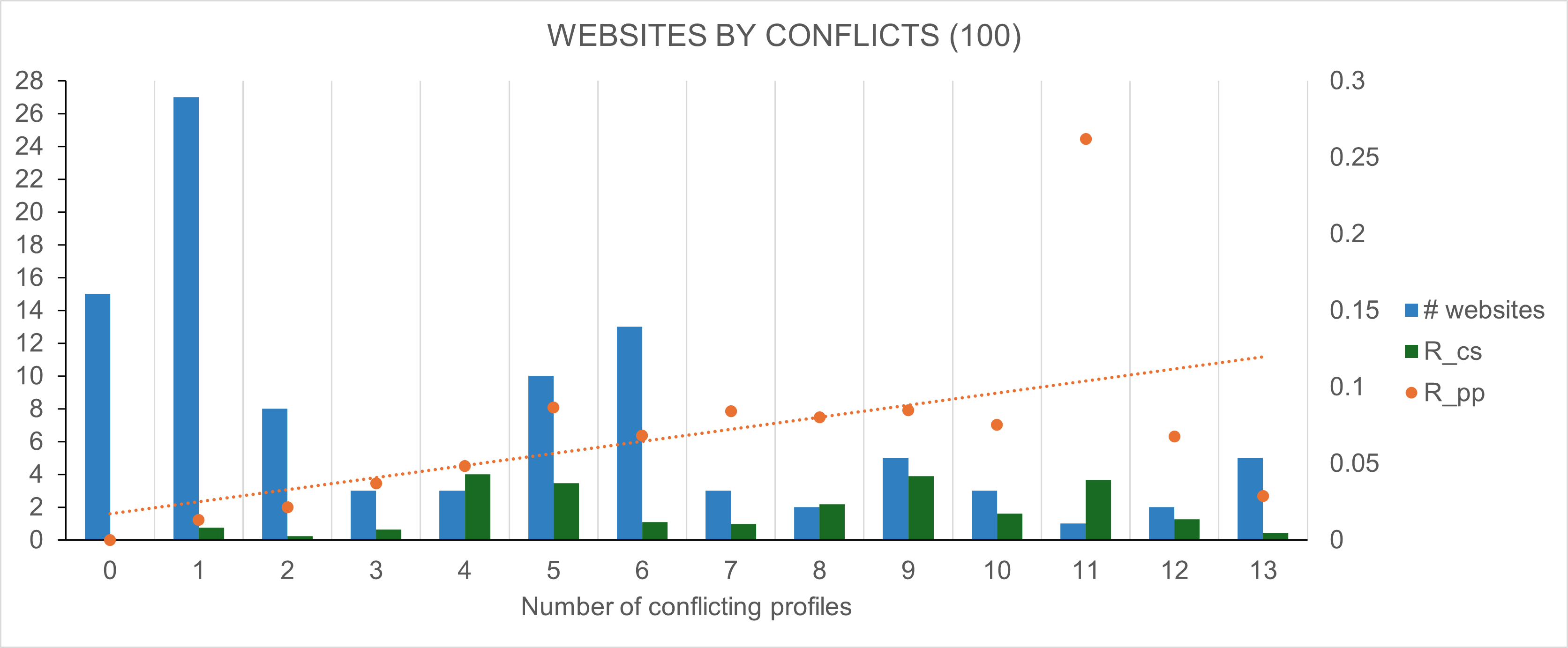}
   \caption{General statistics about conflicts. Dashed orange line is the linear regression model for $R_{pp}$.}
   \label{fig:conflict-stats}
\end{figure}

We first checked if online platforms are compliant with each one of the 23 individual policies, for their respective data types and policies. Figure \ref{fig:conflict-stats} shows the distribution of the number of violated user profiles per website (referred to as each \emph{violation group} hereafter). We see that many websites reside at violation group 0 and 1, indicating that either they did not violate the constructed profiles or the NLP pipeline failed to recognize information for the data practice. The discussions below are all subject to this possibility. However, since we are mainly interested in what conflicts have been detected, this is generally not a concern -- the number would be higher if the NLP pipeline identifies complete information.

We also calculated the average rate of violations per segment, with two variants, $R_{pp}$ and $R_{cs}$, using the following formula (for each \emph{violation group}):
\begin{align*}
R_{pp} & = \frac{1}{|W|} \sum_{w \in W} R^{w}_{pp} \quad &where\quad&R^{w}_{pp} = N^{w}_{con} / N^{w}_{pp} \\
R_{cs} & = \frac{1}{|W|} \sum_{w \in W} R^{w}_{cs} \quad &where\quad&R^{w}_{cs} = N^{w}_{con} / N^{w}_{cs}
\end{align*}
where $W$ denotes the collection of all websites in the \emph{violation group}, $N^{w}_{con}$ denotes the number of conflicts for website $w$, $N^{w}_{pp}$ denotes the number of segments in the privacy policy of website $w$, and $N^{w}_{cs}$ denotes the number of segments that triggers some conflicts of website $w$.

Intuitively, they measure how privacy-respecting (or privacy-violating) the privacy policy is, from different angles. $R_{pp}$ represents how likely a policy segment creates conflicts, which can be further normalized as $R_{pp} / N_{vp}$ where $N_{vp}$ denotes the number of violated profiles (which is a constant in each violation group), measuring how likely a (longer) privacy policy results in (more) conflicts; $R_{cs}$ is the average number of conflicts triggered by a violating segment, measuring how controversial a violating segment is.

As reflected from Figure \ref{fig:conflict-stats}, $R_{pp}$ (orange dots) jitters around a linear regression model (the dashed line) for most of the groups, proportional to the number of conflicting profiles (X-axis). This indicates a similar information density of data practices in their privacy policies. 
The deviation from the regression line at the right end (when the number of conflicting profiles is larger than 11) plausible indicates that the websites have different specificity, especially with more information density.

Looking at the green bars ($R_{cs}$), $vg=4$, $vg=5$, $vg=9$ and $vg=11$ are observably higher than the rest, with $R_{cs} > 2$. This reflects that these websites (on average) creates more than 2 violations per violating segment, indicating each violating segment breaks more profiles, thus being more aggressive in data practice.
In fact, many websites in these groups are associated with social media (\ws{netflix.com}, \ws{facebook.com}, \ws{instagram.com}) and platforming services (\ws{apple.com}, \ws{office.net}), showing clues on the reason.

\begin{figure}
   \centering
   \includegraphics[width=\linewidth]{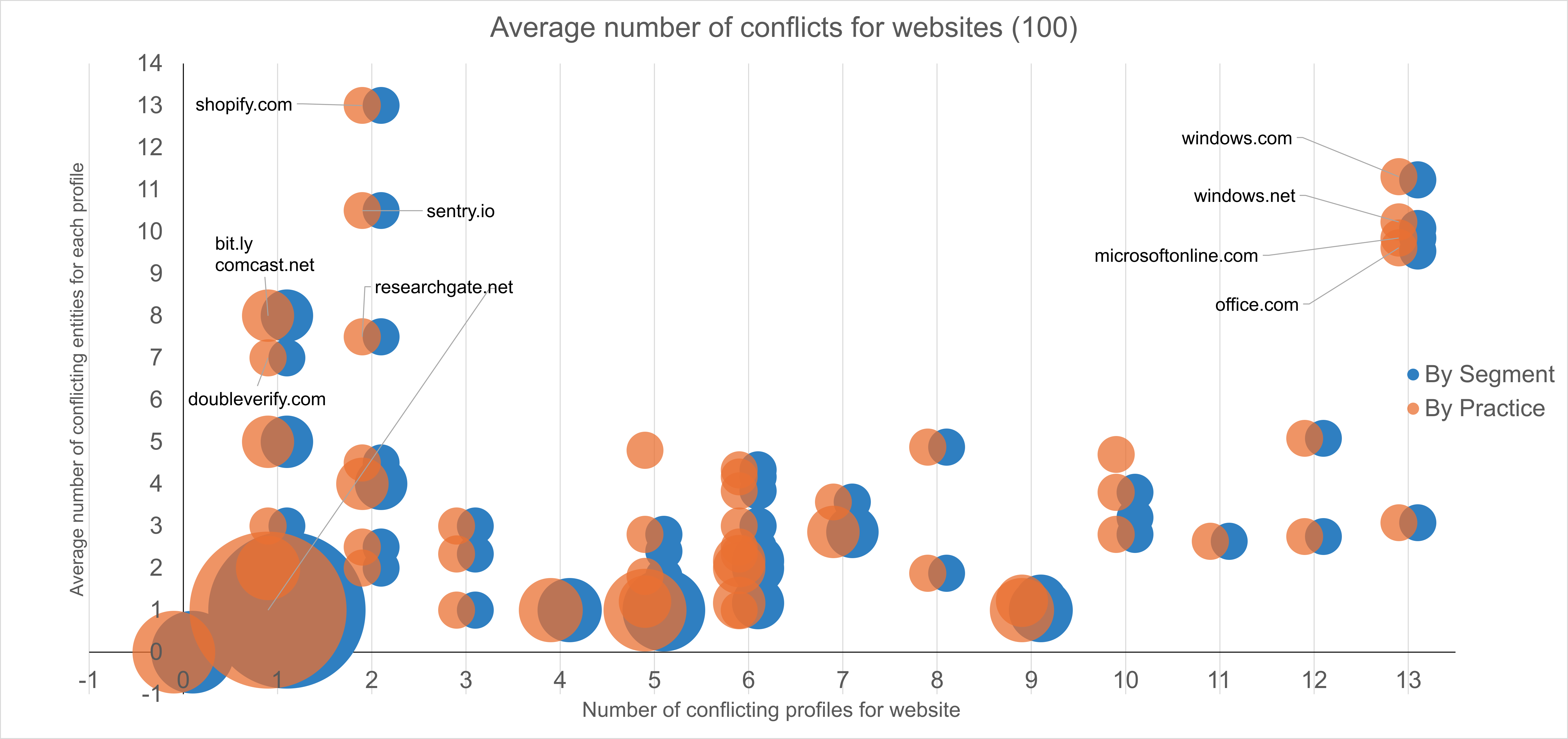}
   % TODO: Add colour based on the type of websites?
   \caption{Conflict distribution of websites. Each bubble represents the websites with the same average number of conflicting entities (by segment or by practice) and number of conflicting profiles.}
   \label{fig:num-conflict-segment}
\end{figure}

Figure \ref{fig:num-conflict-segment} takes a more granule look at the number of conflicting practices:
\begin{equation*}
\frac{1}{N^{w}_{con}} \sum_{p \in P} N_{pr}^{w,p}
\end{equation*}
where $N_{pr}^{w,p}$ refers to the number of conflicting practices for the website $w$ at profile $p$, and $P$ refers to the collection of all profiles\footnote{For completeness, we also show the same measure of number of conflicting segments, instead of practices. Because each (conflicting) segment has around one practice, they are expected to be similar.}. This estimates how controversial a practice can be in the privacy policy. As we can observe, for most websites, the bubbles reside around the 1–5 range, indicating that, despite creating conflicts, they restrain from extensive privacy exploitation. On the other hand, a few websites demonstrate more privacy risks with more than 7 conflicts per profile. Their names are included in the figure, and we can observe that they belong to different business categories, indicating wide-spread privacy risks among Internet services.

For \ws{bit.ly}, \ws{comcast.net} and \ws{doubleverify.com}, only one conflict is identified, the \lstinline|data-ad-3rd-no| profile, which will be discussed later. 
Apart from that, the main source of conflict is related to location data, where only \ws{sentry.io} did not violate that\footnote{After manual investigation, \ws{sentry.io} indeed does not directly collect location data, except for IP address as an indirect source of coarse location.}.
We also observe that \ws{microsoftonline.com}, \ws{office.com}, \ws{windows.net} and \ws{windows.com} all show high number of violations (e.g.~they each consistently violated 6 out of 8 profiles about location data, of 12, 9, 12 and 11 times respectively), but with different numbers, despite belonging to the same company. This is because their privacy policies in the dataset were captured at different time in history, and this result shows that PoliAnalyzer successfully identified their differences, being sensitive to small differences in the policies.

\begin{figure}
   \centering
   \includegraphics[width=\linewidth]{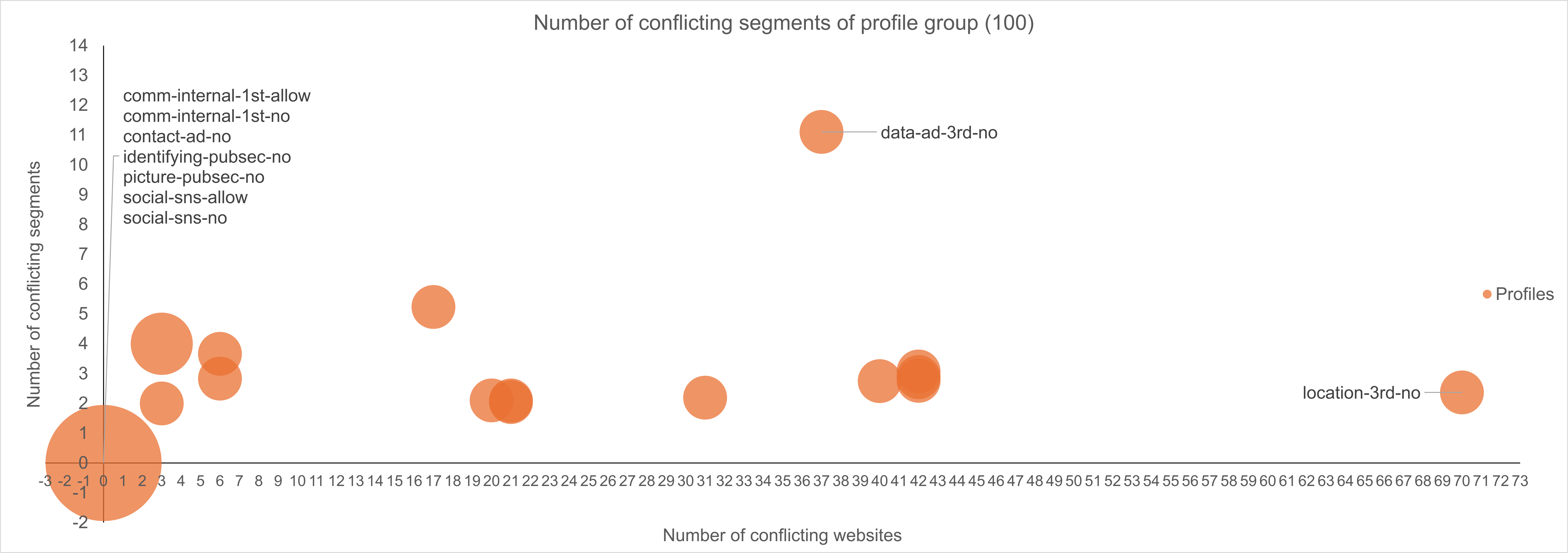}
   \caption{Average number of conflicting segments for profiles. Each bubble represents the profile(s) sharing the same number of conflicting segments and websites.}
   \label{fig:num-conflict-segment-by-persona}
\end{figure}

For an \emph{individual user}, they may be more interested in results from Figure \ref{fig:num-conflict-segment-by-persona}: what types of requirements (profiles) are more likely to cause conflicts, in order to guide their expectations in the online world. In general, the bubbles are sparse on the x-axis (number of websites conflicting with this profile), indicating limited granular details between websites in these user profiles.

There are two exceptional bubbles: \lstinline|data-ad-3rd-no| (on the top) and \lstinline|location-3rd-no| (to the right). Violating \lstinline|data-ad-3rd-no| indicates that many platforms shares data without specifying its type (or the model failed to recognize the type) to 3rd parties for advertisement purposes, with an average of 12 cases for each policy (compared to 1-5 reflected from Figure \ref{fig:num-conflict-segment}). This is a worrying sign that many websites do not clearly detail the data type(s) for 3rd-party sharing for advertisement purposes.

Moreover, violating \lstinline|location-3rd-no| indicates that the location data are shared with 3rd parties for any purposes. The exceptional number of conflicts indicates that the vast majority (70\%) of websites send location data to 3rd parties for processing (either without purposes or purposes unrecognizable by the system), showing a strong signal that location data can be easily exploited. Luckily, there is a glimmer of hope that, when diving into the details of the conflict reports, some segments mentioned the location-sharing practice is optional, either as \emph{opt-in} or \emph{opt-out}, which means the user may still have control over that.

\emph{Auditors} can utilize PoliAnalyzer for different goals, both to gain general understanding of the policies and their compliance with user expectations (from, e.g., Fig. \ref{fig:num-conflict-segment}), or gain more focused understanding on what are the common practices and issues with websites (from, e.g., Fig. \ref{fig:num-conflict-segment-by-persona}). In addition, they can alter or add user profiles to gain more insights for their hypothesis. For example, they may be interested in discovering what location data usage practices are there, apart from those already being examined in these user profiles. For this, they can construct a new profile permitting exactly only the known purposes, for 1st party and 3rd party. We conducted a simulated performance of this task, and discovered an additional purpose \lstinline|RecordManagement|, which is not discussed in literature about user expectations.

Overall, we have shown that different types of users can use PoliAnalyzer for their specific interests and obtain results of different facets, all with only a few number of inputs: in our example, this is 23 (the number of user profiles) plus 100 (the number of websites), and PoliAnalyzer can perform their multiplication of 2300 analysis. User profiles can always be reused for future analysis, thanks for the policy language design. This shows a major methodological improvement compared with existing research (additive vs multiplicative), where each analysis would require one specially constructed input (thus 2300 inputs).

In addition, from our analysis, out of all 100 websites, there are 13205 segments in total, where 3421 segments contain valid practices, and 636 segments demonstrate 4083 conflicts across different profiles. On average, users can achieve a 95.2\% reduction rate if they are only interested in segments creating conflicts, in contrast to all segments, which will be a huge boost for reading privacy policies.

\section{Conclusion}

% Privacy policy is an important but often neglected document for online users, due to the amount of time and focus required to investigate.
In this paper, we presented the design and evaluation of the PoliAnalyzer system, as a means to enable \emph{personalized} \emph{automated} analysis of privacy profiles \emph{at scale}. We use a hybrid approach composed of both neural and symbolic reasoning to achieve scalability, interoperability and auditability: state-of-the-art LLMs are used to identify important data practice information from privacy policies, achieving scalability in documents; the results are converted to logic-based formal representation, as \textit{app policies}, based on the perennial semantic Data Terms of Use (psDToU) language, which are further used in formal reasoning, achieving auditability for both policy source and reasoning; users only need to express their personal preferences using the psDToU language as \textit{data policies}, which is interoperable across different \textit{app policies} for compliance checking. We evaluated the NLP pipeline's performance, using enriched annotations on top of PolicyIE dataset, from domain experts, demonstrating a high overall performance, which is comparable to human annotators. We further synthesized 23 user profiles from existing literature, and use them as requirements to evaluate the top-100 most-visited websites. This both demonstrates the practical usage of PoliAnalyzer to help scalable personalized analysis, and discovers patterns and exceptions in website performance, as well as gaps between existing research on user perceptions and real-world practices.

Overall, the personalized scalable analysis provided by PoliAnalyzer aims to bring attention for online users to regain agency over online privacy practices through concrete discovery, to support decision-making. It can also act as a tool for regulators or activists to understand the distribution of privacy practices of online platforms, and foster targeted conversation, as demonstrated above.
We envision a future where more automation is applied in personalized analysis of privacy and data practices of online activities, and users are freed from focusing on routing actions but more critical ones. PoliAnalyzer acts as a step towards that, but improvements and more work on the similar line should be developed.

We also note the limitations of current research and future directions for PoliAnalyzer, in the next subsection.

\subsection*{Limitation and future direction}

The current work mainly explored and verified whether off-the-shelf LLMs can be used to support information extraction from privacy policies, with minimal effort (e.g.~fine-tuning with a small amount of data). It demonstrated a reasonable performance, but future work may explore more targeted AI models to achieve a better performance, thus improving the accuracy of the system overall, and being potentially more cost-effective.

The types of information can also be expanded, together with more annotations. Related to this, work on improving the formal policy model to support other types of information and reasoning is worthwhile, for more comprehensive analysis with nuanced user preferences.

User ergonomics work may be done in the future as well, such as tools to specify user expectations / \textit{data policies} for non-expert users, or a graphical explanation of the reasoning results. Corresponding user studies to uncover user needs and UX expectations are also worthwhile directions.

\section*{Acknowledgments}

This work is supported by the Ethical Web and Data Architecture in the Age of AI (EWADA) project funded by Oxford Martin School.

\bibliographystyle{named-custom}
\bibliography{rui-main,extra}

\end{document}